# Knowledge Acquisition for Content Selection

Ehud Reiter[1]
Computing Science Dept
Univ of Aberdeen
Aberdeen, UK
ereiter@csd.abdn.ac.uk

Alison Cawsey
Computing and EE Dept
Heriot-Watt University
Edinburgh, UK
alison@cee.hw.ac.uk

Liesl Osman
Medicine and Ther. Dept
Aberdeen University
Aberdeen, UK
l.osman@abdn.ac.uk

Yvonne Roff[2]
Valstar Systems
Burghmuir Drive
Inverurie, UK
yvr@valstar.co.uk

**Abstract:**
An important part of building a natural-language generation (NLG) system is *knowledge acquisition*, that is deciding on the specific schemas, plans, grammar rules, and so forth that should be used in the NLG system. We discuss some experiments we have performed with KA for content-selection rules, in the context of building an NLG system which generates health-related material. These experiments suggest that it is useful to supplement corpus analysis with KA techniques developed for building expert systems, such as structured group discussions and think-aloud protocols. They also raise the point that KA issues may influence architectural design issues, in particular the decision on whether a planning approach is used for content selection. We suspect that in some cases, KA may be easier if other constructive expert-system techniques (such as production rules, or case-based reasoning) are used to determine the content of a generated text.

## 1 Introduction

Knowledge acquisition (KA) is an important aspect of building a natural-language generation (NLG) system, but it has rarely been discussed in the research literature. In this paper we describe our attempts to apply KA techniques developed in the expert-system community to the task of building the part of an NLG system which determines the content and rhetorical structure of a text. This is usually called "text planning" in the literature, but we use the name *content selection* in this paper, to emphasise that this is not necessarily a planning process. We regard content selection as a type of constructive expert-system task [Clancey, 1985]. As such, it may be based on planning, but may also be based on other approaches, such as production rules or case-based reasoning.

Our findings about KA are preliminary, but they do indicate that techniques such as structured group discussions and think-aloud protocols can be very useful in NLG. They also highlight some of the problems with relying solely on corpus analysis or directly asking experts for knowledge, which are probably the most commonly used KA techniques for NLG systems today. Finally, our experiments suggest that in some cases it is difficult to perform KA for planning-based systems, which may be an argument for using other constructive expert-system techniques.

We do not in this paper discuss KA for other NLG tasks (such as realisation and sentence planning), nor do we discuss the use of automated KA or machine-learning techniques. These remain subjects for future research.

In the rest of this introductory section, we briefly describe the project in which this work was carried out. Section 2 gives a brief introduction to knowledge acquisition. Section 3 describes the specific techniques we tried; the successes and failures we observed, and the lessons we have learned about KA in our project. Section 4 discusses the architectural impact of KA on the choice of content-selection technique. Section 5 gives some concluding comments

### 1.1 The Smoking-Letters Project

Our research on KA was done in the context of a project to build an NLG system which generates letters that encourage people to stop smoking. These letters are tailored for each recipient, using information extracted from a questionnaire about smoking attitudes which is filled out by the recipient. The project is loosely based on, and certainly inspired by, the work of Strecher and his colleagues [Strecher et al, 1994]. In particular, the tailoring is partially based on a Stages of Change [Prochaska and DiClemente, 1983] model of behaviour change. We hope to achieve better results than Strecher by using better NLG technology, and also better KA techniques. Figure 1 shows an example of a letter produced by our current pilot system, and Figure 2 shows the questionnaire data which this letter was generated from. The texts produced by the pilot system are not spectacular, and we show an example here primarily to give readers an idea of the inputs and outputs of our system.

Compared to other NLG projects, ours is probably closest to Piglet [Cawsey et al, 1995] and Migraine [Buchanan et al, 1995]. These systems also generated tailored information for medical patients, although with

---
[1] Please address all correspondence to Dr. Reiter at this address.
[2] This work was done while Yvonne Roff was at Aberdeen University, Department of Medicine and Therapeutics

| | | | |
|---|---|---|---|
| **I.** | | Are you planning to quit within the next 6 months? | *No* |
| **II.** | | Are you planning to quit within the next month? | *No* |
| **III.** | | How many cigarettes a day do you smoke? | *20+* |
| **IV.** | | Do you usually have your first cigarette within 30 minutes of awakening? | *Yes* |
| **V.** | | Which of the following are things you like about smoking? | |
| | **A.** | Something to do when bored | *Very Important* |
| | **B.** | It helps me cope with stress | *[blank]* |
| | **C.** | It is something to do with family or friends | *[blank]* |
| **VI.** | | Which of the following are things you *dislike* about smoking? | |
| | **A.** | It is expensive | *Very Important* |
| | **B.** | It is bad for my health | *Very Important* |
| | **C.** | It makes me smell of smoke | *[blank]* |
| **VII.** | | If you decided to try to quit smoking, how confident would you be about succeeding? | *Not confident* |
| **VIII.** | | If you tried to stop smoking, how supportive do you think the following would be | |
| | **A.** | Husband/wife/partner | *[blank]* |
| | **B.** | Other family members | *Not Supportive* |
| | **C.** | Friends | *Not Supportive* |
| **IX.** | | Have you ever tried to quit before? | *Yes* |
| **X.** | | If so, what is the longest time you stopped for? | *1 week to 1 month* |
| **XI.** | | If you have previously tried to quit, why did you go back to smoking? | |
| | **A.** | I put on weight | *[blank]* |
| | **B.** | I felt too stressed | *Very Important* |
| | **C.** | I didn't really want to stop | *[blank]* |
| **XII.** | | Do you have any of the following health problems? If so, do you think smoking contributes to them? | |
| | **A.** | chest pain | *[blank]* |
| | **B.** | cough | *Yes, Yes* |
| | **C.** | wheeze | *Yes, Yes* |
| | **D.** | breathlessness | *[blank]* |
| | **E.** | frequent chest infections | *Yes, Yes* |

*[blank] means no response was checked*

**Figure 2: Extract from Questionnaire from Smoker X**

the perhaps easier goal of increasing compliance (for example, making sure that patients take the medicines they are supposed to take). These systems also used an interactive hypertext interface, whereas our system generates one-off letters. In any case, all three systems attempt to use personalised information to change the behaviour of people with little medical knowledge; this is quite different from, say, NLG systems whose ultimate goal is simply to communicate useful information, such as FoG [Goldberg et al, 1994] and IDAS [Reiter et al, 1995].

## 2 *Knowledge Acquisition*

Knowledge acquisition is acknowledged to be one of the most important aspects of building expert systems, and numerous KA techniques have been developed by the expert-system community [Scott et al, 1991; Buchanan and Wilkins, 1993]. Among other things, this research has shown that it usually is *not* a good idea to ask experts directly for knowledge, because in such cases the knowledge engineer is likely to acquire the textbook knowledge the expert learned in university, rather than the active knowledge that the expert actually uses in his or her day-to-day work. Because the latter knowledge is largely compiled [Anderson, 1995], it is difficult for experts to describe it abstractly using introspection. Usually asking experts to work on specific example cases is the best way of accessing such compiled knowledge [Scott et al, 1991].

### 2.1 Previous Work on Knowledge Acquisition for Content Selection

Surprisingly little attention has been paid to the problem of the KA for NLG systems. In the context of content selection or "text planning", for instance, while numerous papers have been written about architectural and reasoning issues (e.g., schemas vs. planning), very few papers have discussed the problem of acquiring the specific plan operators or schema elements that should be used by the NLG system in the target domain. This is a pity, because we suspect that in NLG (as in other AI systems), good knowledge is as important as good algorithms for the system's performance and usefulness.

[McKeown et al, 1994] briefly describes the KA process used in the PlanDoc system, which we believe is representative of the KA process used to build most current applied NLG systems. In this process, the developers

1. interviewed potential users in order to get a general understanding of the domain and the system's requirements,

2. asked a (single) expert to write a corpus of example reports, and

3. analysed the corpus in various ways.

The final system was "influenced" by this analysis, but did not attempt to directly reproduce the texts in the corpus.

The FoG system [Goldberg et al, 1994] used a variant of the above process where the domain experts (in this case, meteorologists) were allowed to directly write content-selection rules in a special "meteorologist-friendly" pseudo-code. The NLG developers then attempted to map this pseudo-code into real code as directly as possible. The paper acknowledges that direct-mapping limited the flexibility of the system, but this was perceived as an acceptable price to pay for giving domain experts more control over the texts produced by the system.

[Forsythe, 1995] is the most detailed paper we have found on KA issues for NLG. It describes an ethnographically-based set of KA techniques which were used to build the Migraine system [Buchanan et al, 1995]. These techniques included observing doctors and patients in real consultations, and also conducting directed interviews with patients. It is not entirely clear from the papers how much impact Forsythe's work had on the actual implemented Migraine system. Forsythe's work is perhaps closest in spirit to ours, although one important difference is that we are working with a psychological theory (Stages of Change) which predicts which information is likely to be useful in changing people's behaviour. Performing KA in the context of an underlying theory in many ways is significantly harder than performing theory-free KA (see Section 3.1), although we hope it will result in a more effective system.

## 3 KA Techniques in Our Project

During our project, we experimented with the following KA techniques:

1. *Direct acquisition of knowledge:* We asked experts to tell us what should be in the letters.

2. *Creating and analysing a corpus:* We asked experts to write example letters based on specific questionnaires, and then tried to model the rules they used to produce them.

3. *Structured group discussions:* We asked experts as a group to discuss specific questions about letter content.

4. *Think-aloud sessions:* We asked experts to write example letters (as in (2)), but also asked them to "think aloud" as they did so, with their thoughts recorded in a tape recorder.

These are all fairly "standard" techniques from an expert-system perspective, although (at least judging from the research literature) only (1) and (2) have been widely used for building NLG systems.

### 3.1 Directly Asking for Content Knowledge

The most obvious technique for acquiring content knowledge is to directly ask experts to state it. However, as mentioned above, KA experts in the knowledge-based systems community believe that this is unlikely to work well.

This was certainly our experience as well. We made a single attempt to directly acquire content knowledge, when we asked one of our domain experts to describe the structure of letters for people in different stages of the Stages of Change model. We subsequently asked the same expert (along with several others) to write some example letters based on specific questionnaires, and discovered that the specific example letters he (along with every other expert) produced had a different structure from the "general" structure he had initially proposed.

We pointed out this fact to the expert in question, and he subsequently attempted to revise his general structure to more closely conform to the example letters he had actually written; in other words, to combine his "theoretical" and "practitioner" knowledge. This proved to be a difficult and time-consuming task, however. It was relatively straightforward for him to state theoretical knowledge, or to use his practitioner knowledge to produce example letters, but attempting to integrate the two types of knowledge was a research project in its own right, which could not be done quickly or cheaply.

The issue of how to best combine theoretical and practitioner knowledge is an important one for public-health research, and we hope to have more time to think about this in future projects. In any case, however, the goal of our initial KA exercise was to acquire practitioner knowledge, and the above experience agreed with the expert-system finding that directly asking for knowledge is a poor way to get the working knowledge that practitioners use on a day-to-day basis.

**Extract from letter to Smoker X from Expert A**

You already understand the advantages there would be to you if you stopped smoking - better health, more money to spend on other things (more than £1000 more), and proving to yourself that you could do without cigarettes. There are other ways of relieving stress and boredom that you may be able to think of in place of smoking - perhaps a walk outside, or taking up a hobby at home, or anything else you enjoy. And if people and friends don't support you, remember that you are stopping for yourself, not for them, because you want to keep your health and save money

**Extract from letter to Smoker X from Expert B**

You told us that you would like to stop smoking because you feel your smoking is bad for your health, and that it is costing you a lot of money.

Smoking is certainly expensive: 20 cigarettes per day costs £20 per week, which is £80 per month or £1000 per year. There is sure to be something else you could do with that money!

Regarding your health, you experience some breathlessness and coughing and you get a lot of chest infections, and you feel that smoking has a part to play in this. You also suffer from bronchitis, and you are concerned that your smoking may have caused this, and that it will become more serious in the future. All of your concerns are well founded: smoking has been shown to cause all of the symptoms you mentioned, and to increase the risk of developing bronchitis. It has also been shown to increase the risk of getting many other serious illnesses, amongst them heart disease and cancer.

But the good news is, as you know, that as soon as you stop smoking, your health begins to improve and your chances of developing smoking-related illnesses begin to fall back to normal. You would soon notice a considerable improvement in your cough and breathlessness, and you would get fewer chest infections. *It is never too late to stop smoking:* even after many years of smoking, your health will improve if you stop.

**Figure 3: Extracts from 2 Letters to Smoker X from Different Experts**

Accordingly, we did not make any further attempts to directly ask experts for knowledge.

3.2 Creating and Analysing a Corpus

Our next KA effort centred around a "conventional" attempt to create and analyse a corpus. We collected questionnaires from a sample set of about 10 current smokers, and asked 5 experts (one GP, one psychologist, and 3 consultants (specialists)) to write examples of letters which they would produce based on the information in these questionnaires. We did not distribute questionnaires in a controlled way; instead doctors randomly selected as many questionnaires as they felt they had time to look at. This resulted in a total of 11 example letters.

When we obtained our first example letters, we tried to analyse these for content rules (as well as phrasing and stylistic rules) in a way roughly similar to that done in PlanDoc [McKeown et al, 1994]. However, we discovered that although it was possible to derive a more or less plausible set of rules for letters produced by any one expert, it was impossible to derive a consistent set of rules that covered the complete corpus. This is because different experts produced very different letters for the same patients. There were large differences in style (e.g., level of formality) as well as content. We show an example of this in Figure 3, which shows extracts from letters produced by 2 smoking-cessation experts based on the questionnaire shown in Figure 2. Note, for example, that

- Expert A's letter is more informal

- Expert B's letter goes into more detail about cost and health issues

- Expert A includes advice about relieving stress and boredom in the letter body (Expert B attached a separate advice sheet to his letter)

- Expert B introduces information about a health risk (cancer) which X did not indicate that she was concerned about (see also the "think-aloud" extract shown in Figure 4).

This example is typical of the variation seen, and in no sense extreme.

In short, it proved extremely difficult to conduct a straightforward corpus analysis when the corpus contained texts produced by several experts. Note that in PlanDoc, the corpus analysis used texts from a *single* expert. It would be very interesting to see if similar problems would arise in this domain if a multi-expert corpus was used.

On a more practical note, we now believe it was a mistake to randomly select questionnaires that served as the basis of the example letters. In future KA attempts, we plan to exercise much more control over the selection of example questionnaires, so that they span as much as possible of the space of possible system inputs. We have also discussed "faking" questionnaires for important boundary cases, if these boundary cases do not occur in the data we collect. Correctly handling boundary cases is one of the most difficult aspects of NLG (indeed, of any software system), and the corpus should contain letters showing appropriate letters for boundary cases.

3.3 Structured Group Discussions

After we built the initial corpus and discovered the large variations between experts, we then arranged to have some of the experts discuss these variations as a group. 3 of the 5 experts (one GP, one psychologist, and one consultant) participated in these discussions. The other 2 experts did not participate, partially because of time constraints (which highlights the fact that good KA requires substantial amount of time from experts). As far as possible, we attempted to focus the discussions on specific issues; in particular, we did *not* ask experts to judge which individual letters were best. Following advice from a local KA expert (Prof. Derek Sleeman of the Aberdeen CS Department), we presented these issues as binary choices between two alternatives. In cases where a range of possibilities existed, we listed two "plausible" points in our write-up, but encouraged the experts in the discussion to think of other possibilities as well.

For example, one variation seen in the letters was whether they acknowledged that smoking had benefits as well as costs, or just presented negative information about smoking. We presented this issue to the experts as a binary question in this form ("should letters acknowledge that smoking has good effects as well as bad effects"). The experts decided that letters aimed at some types of patients ("precontemplators" and "contemplators" in the Stages of Change model) should indeed explicitly acknowledge that smoking can have good effects (essentially so recipients perceived the letters as balanced). They also decided, however, that only a sentence or two should be devoted to this, and the rest of the letter should stress the negative aspects of smoking. In other words, we obtained from this discussion not just a binary decision ("include" instead of "don't include"), but also fairly detailed information about when to include this content, and at what level of detail.

Somewhat to our surprise, incidentally, we discovered that in all cases agreement was reached, usually fairly quickly. We were particularly gratified to see that individual experts did not stick to the style they adopted in their letters, but were willing to openly discuss the alternatives.

KA experiences in the KBS community suggest that sometimes discussions are best if they focus on specific examples instead of abstract questions [Scott et al, 1991]. We did not do this here, partially because we were worried that experts might feel defensive if specific letters which they produced were being discussed. In future KA exercises, however, we may experiment with this approach.

3.4 Think-Aloud Sessions

The final technique we experimented with was asking experts to "think aloud" into a Dictaphone while they wrote example letters. As mentioned above, one of the problems with pure corpus analysis is that there is no information on the reasoning or intentions of the document writers. We hoped that asking experts to explain what they were doing, as they worked on an example task, would give us knowledge about their underlying reasoning and intentions. Think-aloud protocols are again a standard expert-system KA technique.

The think-aloud sessions certainly showed that much more reasoning was going on than was evident by simply analysing the example letters as a corpus. For example, the sessions revealed that experts often performed a preliminary classification and analysis of patients before starting to write the actual letters; this was not something we picked up from the corpus analysis. Unfortunately, some of the reasoning described was probably too complex to be implemented on a computer. One expert, for example, talked about achieving the goal of "establishing empathy with a patient" (this may be related to the "enlistment" seen by Forsythe [1995]); it is difficult to see how to encode such a goal in a computer NLG system.

An extract from one of the think-aloud transcripts is shown in Figure 4 (note that the transcriptionist did not record pauses or content-free phrases such as "Um"). This comes from the session where Expert B wrote the letter which is partially shown in Figure 3. *Section Four* refers to the last two paragraphs shown in Figure 3.

The think-aloud sessions were very interesting and thought-provoking (for the experts taking part in them as well as the NLG system developers), and showed how much reasoning was being performed by experts when they wrote a letter. It may be impractical to build a computer system which can reason in as much depth.

I am just thinking through what the main issues are for this person, and I am also thinking through this whole issue of what is important for someone at pre-contemplation stage. Certainly, I think we need to reinforce her perceptions of the bad things about smoking, and that is expense and the effects on her health. And certainly pick up on her particular health concerns. And I would certainly want to mention what she sees as the good things about smoking, namely that it helps her boredom and stress levels. It is a question of how specific you would want to be regarding this at the pre-contemplation stage, it may be just a matter of raising the issues that there may be other ways of dealing with boredom or stress other than smoking; or it might be appropriate to go on and give specific information again...

So, coming to the actual letter...

*Section Four*: picking up here on her specific health concerns, to the fact that she does currently attribute cough, wheeze, and frequent chest infections to her smoking. And she does have bronchitis, which, again, she attributes to smoking, and she is concerned about bronchitis in the future. So, specifically, I would mention the three particular symptoms which she is experiencing and I have mentioned the bronchitis issue particularly as well. I did note that her only health concerns seem to be around the issue of chest problems, and I have put in a sentence just to state that smoking also causes other health problems such as heart disease or cancer. I think there is certainly a place here for confirming her perception that stopping smoking would improve her health.

**Figure 4: Extract from Think-Aloud Session (Expert B on Smoker X)**

3.5 KA and the Development Process

Because of time and resource limitations, we were only able to make a single pass at KA before starting to develop our prototype letter-generation system. This caused a number of problems, because additional issues often arose when we analysed the KA data and started writing code. For example,

- The think-aloud sessions showed substantial differences in how the Stages of Change model was interpreted by different experts. For example, some but not all experts gave advice on techniques for quitting to people who had not yet decided to quit (i.e., rated as Contemplators or Precontemplators in the Stages of Change model); this was something we would have liked to discuss in further KA sessions.

- We had decided at a group discussion session to limit core letters to one page (possibly with another page of attachments), and discussed what material should be eliminated to achieve this size bound. Code development, however, highlighted the fact that we also had a problem with letters that had too *little* content, where there simply wasn't enough to say to fill a page. Presumably general smoking-cessation information could be inserted into such letters; this again was an area where additional KA sessions would have been very useful.

In future projects, we plan to use an iterative development strategy, with KA activities being scheduled throughout the life of the project. This is a standard practice in expert-system development.

3.6 Summary of Lessons Learned

Our experiments were obviously very preliminary, but we think they have revealed some potentially valuable insights for KA in NLG, including

- Directly asking experts for knowledge often doesn't work, its usually better to observe experts performing a task. This agrees with findings about KA in the KBS community.

- A corpus-based KA approach, which seems to be the most common approach used in previous NLG projects, may cause problems because (a) experts may not be consistent with each other, and (b) no information is obtained about the reasoning and intentions of the experts.

- Group discussions of specific questions worked surprisingly well, agreement was usually reached very quickly. We may have been lucky in that our experts were a very co-operative and open-minded group, it would be interesting to try discussions with other groups of experts.

- Think-aloud sessions provided good information on reasoning and intentions.

- KA sessions should be part of an iterative development strategy, as is done in most expert-system projects. A good KA session will raise almost as many questions as it answers, and some issues will only surface once code is being developed.

- KA is time-consuming (and therefore expensive), and requires considerable amounts of time from the domain expert as well as the

developers. This needs to accepted and incorporated into the project plan.

## 4 *Architectural Implications*

As mentioned in the introduction, we regard content selection (or "text planning") as a type of constructive expert system. As such, in principle there are many ways in which it can be performed, including planning, schemas, production rules, case-based reasoning, and so forth. One influence on the choice of algorithm which perhaps has not received much attention to date is KA. In particular, an approach which requires a substantial amount of difficult KA may be less appealing than an approach which requires relatively simple KA.

More specifically, KA difficulties may explain why planning techniques have not been used in many applied NLG systems, despite the many advantages of planning which have been pointed out by Hovy [1988], Moore [1993], and others. While some applied NLG systems have used planning-based representations, to the best of our knowledge very few (if any) have used planning in the way described by Hovy [1990, page 17], which is "identifying the basic building elements from which coherent paragraphs are composed and developing a method of assembling them dynamically into paragraphs on demand". Those applied NLG systems which have used planning-like representations (such as Migraine [Buchanan et al, 1995]) have instead tended to simply expand top-level "goals" using hierarchical decomposition. In other words, the planning notation has been used to represent what amounts to a set of macros which are expanded to produce the text structure.

We believe that one reason why planning has not been used more in practice is that if done in the sense described by Hovy, it requires a large amount of precise and exact knowledge about how intentions are satisfied in the domain. It can be very difficult and expensive to acquire such knowledge, especially in health domains, where there is no rigorous underlying theory which can be used to motivate reasoning. It is often significantly easier to acquire the more "heuristic" knowledge seen in schemas, production rules, or case-based reasoning; and this in itself may be a good argument in many applications for using these techniques, even if in principle planning would lead to a more robust and powerful system.

We in no sense are claiming that planning will *never* be used for content selection in applied NLG systems, but rather that planning is simply one of a set of possible techniques for this task, which has numerous advantages (well described by Hovy and Moore) but also some disadvantages. In domains where a strong underlying theory makes KA easier, planning may be justified. But in domains where KA for planning would be very expensive, it may make more sense to use other techniques.

### 4.1. Knowledge Acquisition for Planning

One aspect of many KA exercises is that much of the knowledge they acquire is compiled, and that information about intentions and underlying reasoning is often not given (at least initially). Certainly no information about intentions and reasoning is obtained in a pure corpus analysis, which (as mentioned in Section 2.1) is probably the dominant form of KA in most current applied NLG efforts. In other words, a corpus analysis tells the developer *what* the expert did but not *why*. Hence it can be used to develop systems based on schemas, production-rules, or case-based reasoning, but it cannot be used to develop a planning system which reasons about intentions and ways of fulfilling goals.

KA for a planning system must then be heavily based on techniques such as structured discussions and think-aloud protocols, since these do provide some insight into reasoning, intentions, and goals. However, even here the information provided is usually not complete. For instance, in the last paragraph of the think-aloud protocol shown in Figure 4, the expert asserts that he would mention in the letter the symptoms and health problems which the patient attributes to smoking; but he does not say whether he would simply repeat these concerns, or go into more detail (for example, "X% of smokers suffer from bronchitis, but only Y% of non-smokers have this problem"). If we examine the actual letter written by the expert (shown in Part B of Figure 3), we can see that in this case the expert has not in fact gone into any additional detail. In contrast, the expert did go into additional detail earlier in the letter when discussing the expense of smoking, by calculating a specific yearly cost.

If we were building a system based on schemas or production rules and consistently saw this pattern in the corpus, we could simply encode a rule that the system should give figures for the cost of smoking but not for smoking-related health risks. If we are building a proper intention-based planning system, however, this approach will not suffice, and we will need to question the expert to determine why he gives figures for cost but not health risks. Doing this should of course in the end produce a more robust and useful system; but it will also require a non-trivial amount of KA.

This case is by no means atypical. Even though we have in our project obtained think-aloud protocols and organised structured discussions, there are still many, many, areas where we can observe rules but not the underlying rationale. This should not be a surprise, since human experts heavily rely on "compiled knowledge" [Anderson 1995]. Of course we can schedule additional KA sessions to attempt to understand the rationale behind these rules, and this should in principle lead to better letters. From an engineering cost-benefit perspective [Reiter and Mellish, 1993], however, it makes sense to do this selectively, in cases where we think understanding the rationale would significantly improve the letters; insisting on acquiring the reasoning/intentions behind *every* rule would be extremely expensive.

In other words, the expense of acquiring information about intentions and reasoning means that from a practical perspective, it is probably best to mix "deep" plan-based reasoning with shallower reasoning based on observed rules and heuristics.

### 4.2. Meeting Ill-Defined Goals without a Rigorous Domain Theory

A related issue which we wish to raise is that planning may be more difficult in domains, such as health, where goals are often ill-defined and there is no strong underlying domain theory. In areas such as logistics, there are clear goals (such as "move 2000 widgets from X to Y") and a good underlying theory (for example, we know that ship A can carry 1000 widgets, and will take between 4 and 5 days to get from X to Y). This means KA activities in these domains can concentrate on high-level rules. There is less need for detailed KA, because the domain theory can provide a rationale for well-specified subgoals. For example, if ship A has a capacity of 1000 widgets, we do not need to run KA sessions to determine why experts load ship A with 1000 widgets instead of 1100 widgets or 900 widgets; the domain theory explains this.

In health domains, in contrast, goals are fuzzier and there is no mathematically rigorous domain theory. The smoking-letters project probably has an unusually clear overall goal (get the patient to stop smoking), at least compared to goals such as "reduce anxiety" and "improve information exchange" found in Piglet [Cawsey et al, 1995] and Migraine [Buchanan et al, 1995]. But many of the sub-goals that appeared in our KA sessions, such as "build up empathy", are poorly defined. We are also fortunate in the smoking project to have a psychological theory (Stages of Change) which gives some guidance as to how our top-level goal can be achieved [Prochaska and Goldstein, 1991]; again, Piglet and Migraine did not have such theories. But Stages of Change only gives very broad guidance, and indeed our attempt to build a detailed letter model based on this theory was not successful (see Section 3.1).

Another problem is that in some health domains, there may be many goals, whose interaction is poorly understood. For example, [Cawsey and Grasso 1996] point out that when experts discuss how to explain potential side-effects to patients, they have many goals, including helping patients cope with side effects, reducing anxiety, and not overloading patients with information. Furthermore, these goals can conflict. For example, telling a patient how to cope with an extremely unpleasant but unusual side-effect helps in achieving the coping goal, but may detract from the anxiety-reduction goal. As Cawsey and Grasso point out, it is difficult to incorporate to program a planning system to deal with such goal conflicts, especially given our poor understanding of how to achieve these goals individually.

In other words, the planning approach may be especially difficult to follow in domains (such as health and medicine) which have ill-defined goals or subgoals, lack a mathematically precise domain theory, and/or often have conflicting goals. These factors mean that substantially more KA will be required to build a planning-based system than would be needed in a domain (such as logistics) with precise goals and a rigorous domain theory.

### 4.3 Techniques for Content Selection

The NLG research literature has concentrated on planning and schemas as techniques for content selection. But there are many other ways of performing this task, including

*Case based reasoning*: CBR involves retrieving past cases which "nearly" match the current situation, and adopting them to be appropriate for the current context. This approach is widely used in business-letter applications based on mail-merge technology (Coch, 1996), and could be adopted to NLG, perhaps using some of the document repair techniques described in [DiMarco et al 1995]. KA for CBR requires collecting and classifying a set of example letters, and also devising a set of adaptation rules.

*Production rules*: Rule-based systems use sets of if-then rules to specify when a particular piece of information should be included in a text. One NLG world which used rules for content selection was IDAS [Reiter et al, 1995], and indeed this decision was partially justified on the grounds that it made KA easier [Reiter and Mellish, 1993].

*Propose and Revise:* Propose and revise [Jackson 1990] is a technique where an initial design (based on a rough model of what is required) is pruned and extended using more specialised knowledge and constraints. We do not believe propose-and-revise has been used for content selection in the past, but one of us (in a different project) is currently investigating using this technique for content selection, partially because it seems well-suited to meeting ill-defined goals (Section 4.2).

The above lists only a few of the possible alternatives to planning and schemas for content selection. There are many other ideas and techniques which have been developed by the constructive expert-system community (for example, see [Jackson 1990]), often with easy KA as a motivating factor, which we believe could be used for content selection.

## 5 Conclusion

In conclusion, we have tried to make two basic points in this paper. Firstly, we have argued that knowledge acquisition for NLG systems should not be restricted just to corpus analysis and directly asking domain experts for rules. Instead, NLG developers should consider using some of the many other KA techniques developed by the expert-system community, such as think-aloud protocols and structured group discussions. Secondly, we believe that KA has an impact on the choice of NLG architectures. In particular, because acquiring intentional information can be very expensive in KA terms (at least in domains such as health which lack a precise domain theory), it may be difficult to use planning for content selection in some cases, and better to use other techniques, such as schemas, rules, or case-based reasoning.

**Acknowledgements**

We would like to thank all of the experts who took part in our KA sessions, especially Dr. Scott Lennox and Dr. James Friend. We also greatly appreciate Prof. Derek Sleeman's advice on KA procedures. Much of this work was funded by the Scottish Office, Home and Health Department.

**Figure 1: Example Output Letter**

<div style="text-align: right">
Chest Clinic<br>
City Hospital<br>
ABERDEEN<br>
AB24 5AU
</div>

Dear X

Thank you for completing the smoking survey recently. From your answers it sounds as though you are quite a heavy smoker who is experiencing some chest problems  I see that you have managed to stop smoking in the past but stress  made you start again.

Although you said that you did not intend stopping in the foreseeable future, I hope some of the following information might be of interest to you.

## Most People Who Really Want To Stop Can, And Do Stop.

When smokers have a cigarette, one of the pleasures is having another dose or 'fix' of nicotine which your body has come to depend on.  The problem is that along with the nicotine comes all the harmful things which give you your coughing, wheezing, chest infections and bronchitis and which may eventually make you permanently breathless from lung damage.

You already understand the advantages there would be to you if you stopped smoking -  better health,  more money to spend on other things (more than £1000 a year more),  becoming fitter, proving to yourself that you could do without cigarettes.

If you do decide to stop then many people have found the following method useful. Choose a day to stop, and then try not to smoke when you wake in the morning for the first half-hour. At the end of this, see if you can manage another half-hour - don't look too far ahead - and see if you can keep going a bit longer, and before you know it, you may have got through to dinner-time.  Keep on delaying and delaying, and keep busy with other things to try and take your mind off it. It won't be easy, but it will get easier as the days go by, and the craving will gradually get less. If you don't succeed, just pick another day and try again. The most important factor in stopping is really wanting to.

I hope this letter has given you some food for thought, and that it might help you to try and stop smoking in the future.

Best Wishes

 A. Doctor

---

*Facts About Smoking*

Around 5,000 British people die in road accidents each year - smoking kills 20 times as many.